\documentclass[10pt,twocolumn,letterpaper]{article}

\usepackage{cvpr}
\usepackage{times}
\usepackage{epsfig}
\usepackage{graphicx}
\usepackage{amsmath}
\usepackage{amssymb}
\usepackage{enumitem}

\usepackage[utf8]{inputenc}
\usepackage{natbib}
\usepackage{xcolor}
\usepackage{url}
\usepackage{booktabs}
\usepackage{graphicx}

\usepackage[breaklinks=true,bookmarks=false]{hyperref}

\cvprfinalcopy 

\def\cvprPaperID{****} 

\title{Retouchdown: Adding Touchdown to StreetLearn as a Shareable Resource for Language Grounding Tasks in Street View}
\author{
Harsh Mehta \\
Google Research \\
{\tt\small{harshm@google.com}} \\
\and
Yoav Artzi \\
Cornell University \\
{\tt  \small{yoav@cs.cornell.edu}} \\
\and
Jason Baldridge \\
Google Research \\
{\tt  \small{jasonbaldridge@google.com}} \\
\and
Eugene Ie \\
Google Research \\
{\tt  \small{eugeneie@google.com}} \\
\and
Piotr Mirowski \\
Deepmind \\
{\tt  \small{piotrmirowski@google.com}} \\
}
\date{December 2019}

\begin{document}

\maketitle

\begin{abstract}
The Touchdown dataset \citep{Chen19:touchdown} provides instructions by human annotators for navigation through New York City streets and for resolving spatial descriptions at a given location. To enable the wider research community to work effectively with the Touchdown tasks, we are publicly releasing the 29k raw Street View panoramas needed for Touchdown.  We follow the process used for the StreetLearn data release \citep{streetlearn} to check panoramas for personally identifiable information and blur them as necessary. These have been added to the StreetLearn dataset and can be obtained via the same process as used previously for StreetLearn. We also provide a reference implementation for both of the Touchdown tasks: vision and language navigation (VLN) and spatial description resolution (SDR). We compare our model results to those given in \cite{Chen19:touchdown} and show that the panoramas we have added to StreetLearn fully support both Touchdown tasks and can be used effectively for further research and comparison. 
\end{abstract}

\section{Introduction}

Following natural language navigation instructions in visual environments requires addressing multiple challenges in dynamic, continuously changing environments, including language understanding, object recognition, grounding and spatial reasoning.  
Until recently, the most commonly studied domains were map-based~\citep{Thompson:93maptask} or  game-like~\citep{Macmahon06walkthe,Misra:17instructions,misra-etal-2018-mapping,hermann2017grounded,hill2017understanding}. 
These environments enabled substantial progress, but the complexity and diversity of the visual input they provide is limited. This greatly simplifies both the language and vision challenges. 
To address this, recent tasks based on simulated environments include photo-realistic visual input, such as Room-to-Room~\citep[R2R;][]{Anderson:2018:VLN}, Talk-the-Walk~\citep{talk_the_walk} and Touchdown~\citep{Chen19:touchdown}, all of which rely on panorama photos. 

A major challenge of creating simulations that use real-world photographs is they at times capture bystanders and their property.
This raises privacy concerns and requires additional care to check for and ensure personally identifiable information (PII) is removed from research resources that are made publicly available. 
Existing resources adopt different strategies to address this. 
The Matterport3D dataset~\citep{Matterport3D}, which underlies the R2R task, is focused on real-estate data that is curated to exclude PII. This approach is limited to environments of a specific type: houses that are for sale. Academic resources that focus on urban street scenes opted to manually collect panoramas from scratch and scrub them for PII \citep{talk_the_walk,Weiss2019:sevn}. This is laborious and costly---especially the first stage of collecting the panoramas. As a result, such resources cover relatively small areas.

Google Street View has world-wide scale coverage of street scenes. Each panorama in Street View has gone through a process to protect the privacy of bystanders and their property. Individuals can also request specific panoramas to be removed. As such, it is a resource with the potential to transform the research community's ability to study problems such as street scene understanding and navigation. Touchdown relies on 29,641 panoramas from Street View; however, because raw images cannot be distributed according to the Street View terms-of-service,\footnote{\url{https://www.google.com/help/terms_maps/}} these are not provided with the Touchdown data. Instead, only image feature vectors are available for direct download with the data, and access to the raw panoramas is subject to availability through APIs governed by Street View's terms of service. 

Research can be done within a company and shared via publication without releasing data; for example, \cite{cirik2018following} discussed models for instruction-conditioned navigation in Street View. However, the full impact of the data and research about it can be better realized by making at least some portion of such resources available to the broader research community.  In this context, StreetLearn~\citep{mirowski2018learning,streetlearn} stands out as a publicly available resource of Street View data that has been approved for dissemination and use for academic research.\footnote{\url{http://streetlearn.cc}} 
StreetLearn contains 114k panoramas from New York City and Pittsburgh that have been manually checked for PII, ensuring, for example, that faces and license plates are blurred. 
The dataset can be easily accessed. 
Researchers interested in working with the data simply fill a form stating their goals and commit to update the data periodically with newer versions as they are released. 
This process balances the ability of researchers to use the data with preserving the privacy and rights of individuals impacted by the data. 
For example, periodic updates allow Google to respond to user takedown requests. 

To increase the accessibility of Touchdown and providing an example of how important data can be responsibly released, we integrate the Touchdown task and its corresponding Street View data into an updated version of StreetLearn. This paper reconciles Touchdown's mode of dissemination with StreetLearn's, which was designed to  adhere to the rights of Google and individuals while also simplifying access for researchers and improving reproducibility. We also provide open source implementations of both the vision-and-language navigation and spatial description resolution tasks, which we show to have a consistent performance with the results in the original Touchdown paper. We hope that this release of data and code will enable the entire research community to make further progress on these problems and to consider new questions and tasks enabled by this limited but significant slice of Street View data. 
\section{Process}
\label{sec:process}

Touchdown includes tasks for natural language navigation and spatial reasoning in realistic urban environments. 
Touchdown uses Street View panoramas of New York City to define a large-scale navigation environment. 
It includes 9,326 human-written instructions and 27,575 spatial description resolution tasks. 
Touchdown's instructions were written by people and emphasize attributes of the visual environment as navigational cues. This makes Touchdown  a valuable resource for research on following natural language instructions in visual environments. This contrasts  with the template-based navigation instructions used by \cite{hermann2019learning}, which were generated by Google Maps API and used with StreetLearn panoramas.

\begin{figure}
  \centering   
  \includegraphics[scale=.25]{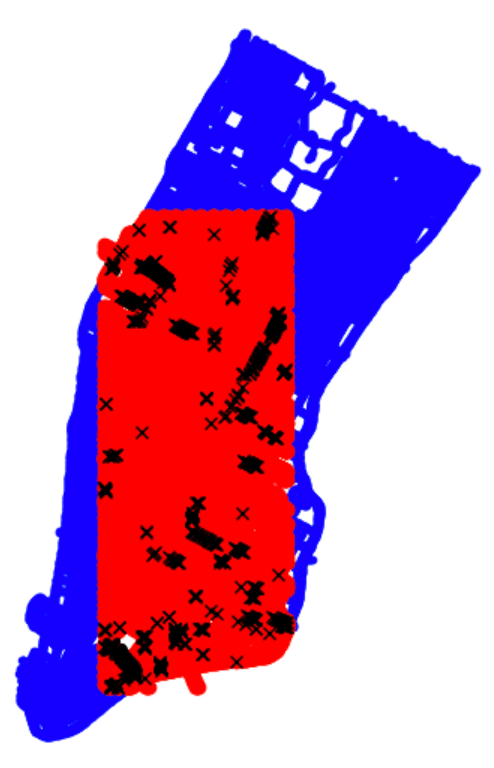}

\caption{ 
    The overlap between the StreetLearn (blue) and Touchdown (red) panoramas in Manhattan. There are 710 panoramas (out of 29k) that share the same ID in both datasets (in black).
    }
    \label{fig:td_sl_overlay}
\end{figure}

\begin{figure*}
  \centering   
  \includegraphics[width=0.9\textwidth]{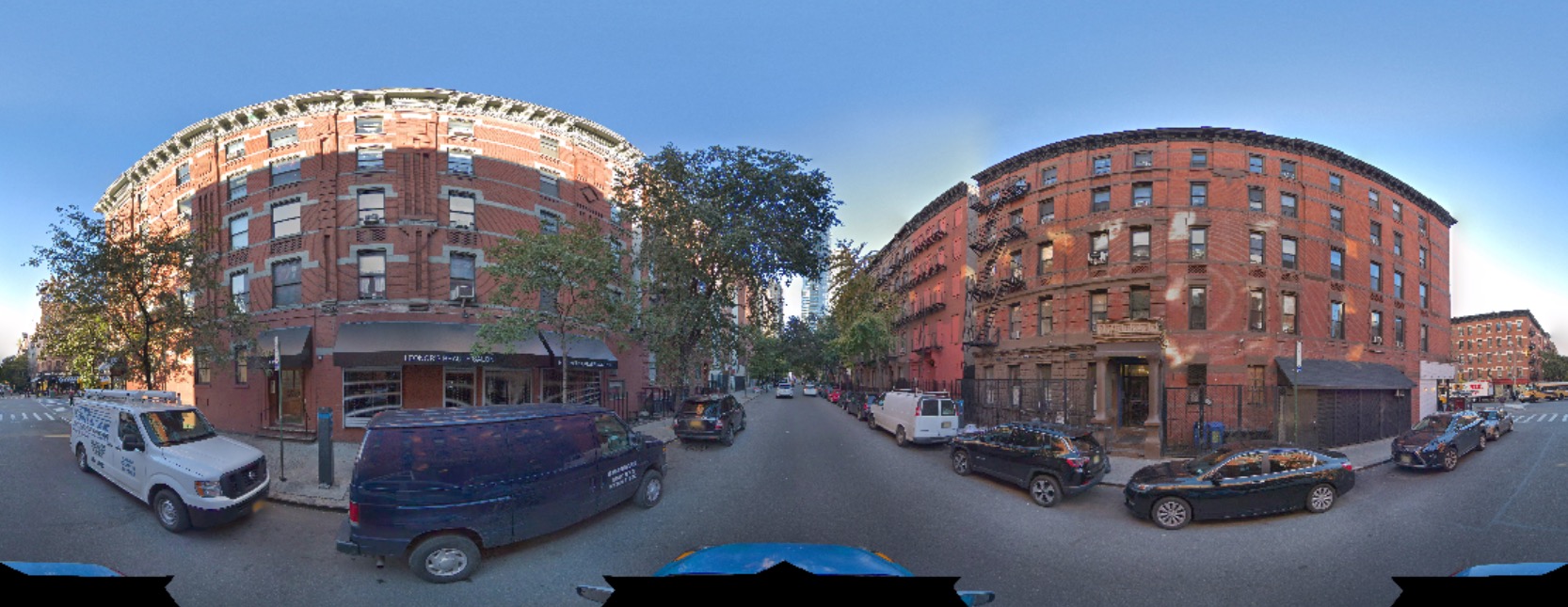}

\caption{ 
    One of the panoramas taken from the dataset which shows transient objects being referenced in the navigation text.
    \emph{``Stop here, and turn left. You will now be walking down a narrow lane with parked \textbf{cars} on both sides. There should be a payphone on your right and a fire hydrant (behind silver poles) on your left. Walk down this lane, and on your left you will soon see a shop with gray columns between the windows and a blue sign with yellow trim."}
    }
    \label{fig:td_car_pano}
\end{figure*}

Unfortunately, the development and release of Touchdown introduced several challenges that complicate working with the data. 
Even though Touchdown itself does not  contain Street View data, it references specific Street View panoramas and depends on access to them via the Street View API. 
This requires any researcher that wishes to work on the data to download large amounts of data using the API, which is inconvenient, error-prone and not aligned with the current Google Maps terms-of-service. 
Also, the panoramas available through the API periodically change, potentially making parts of the data unavailable. This means there is no hope for consistent versioning (which hurts reproducibility) regarding panorama availability because the data collected by each researcher is dependent on the particular time they access it. 
Finally, individual researchers or research groups cannot themselves comply with takedown requests---a responsibility that should stay with Google. Therefore, long term storage must be kept within Google, with researchers periodically refreshing the data.



\begin{figure*}
    \centering
    \textbf{Panorama before the main SDR panorama.}\\
        \includegraphics[width=.9\textwidth]{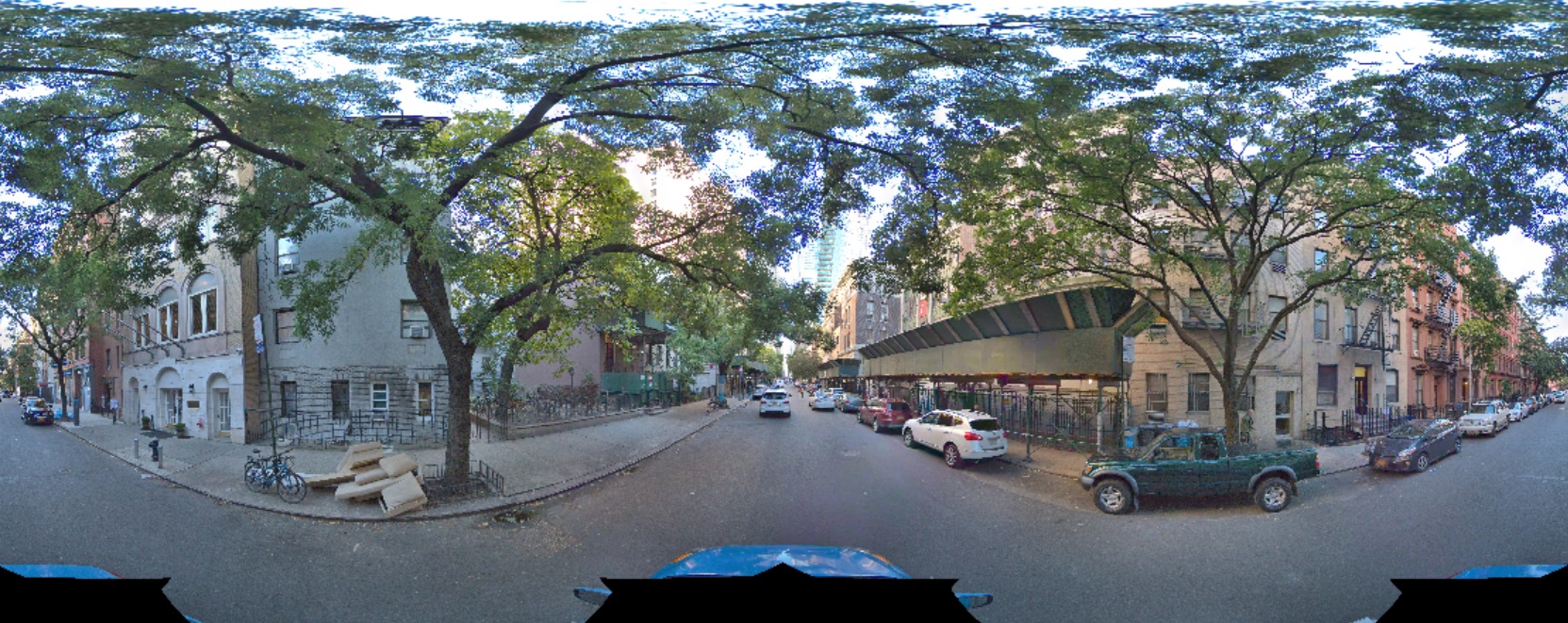}
    \\\vspace{.5cm}
    ~ 
    \textbf{Main SDR Panorama.}\\\vspace{.1cm}
        \includegraphics[width=.9\textwidth]{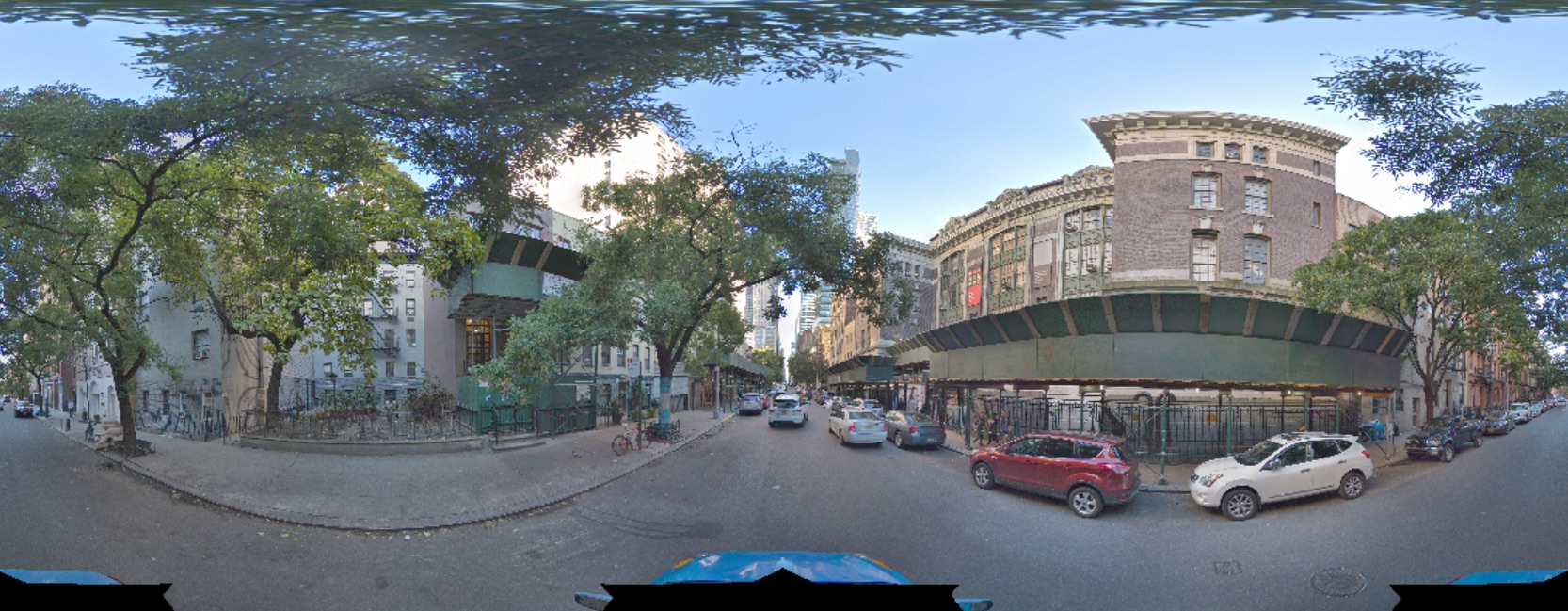}
    \\\vspace{.5cm}
    ~ 
    \textbf{Panorama after the main SDR panorama.}\\\vspace{.1cm}
        \includegraphics[width=.9\textwidth]{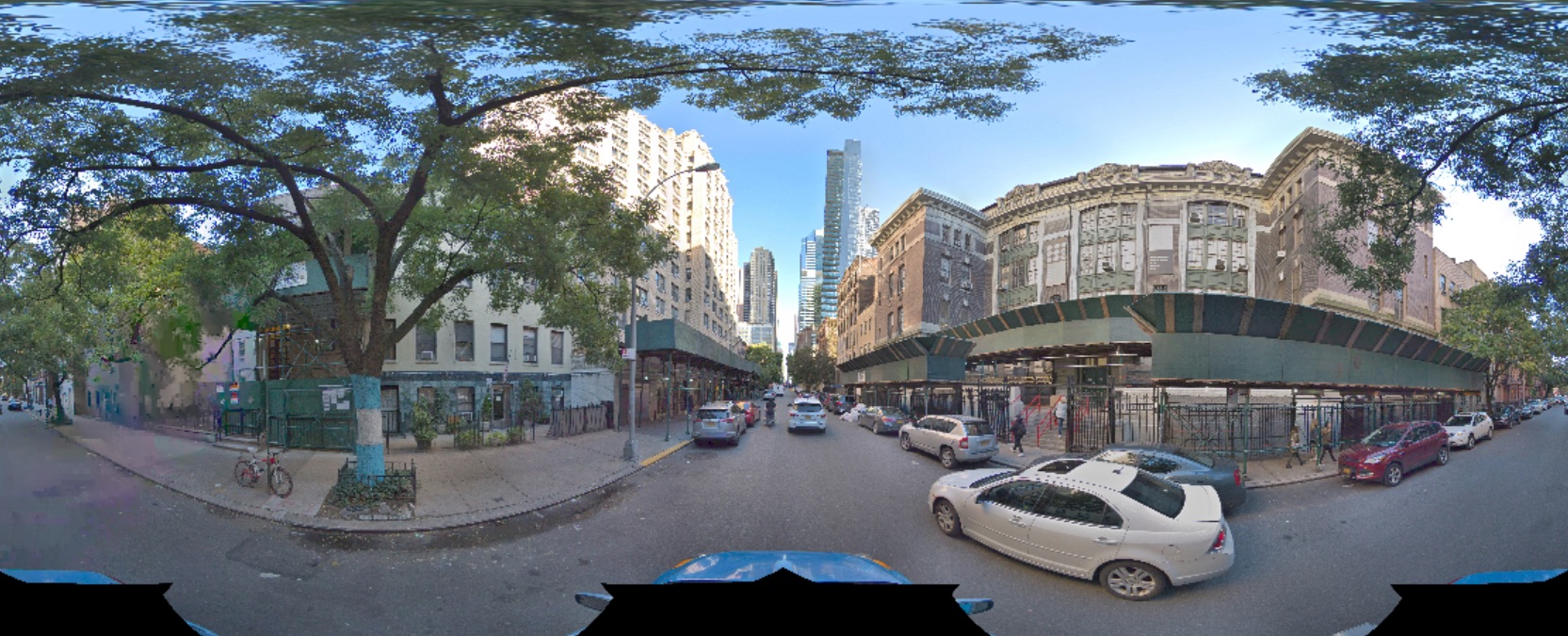}
    \\\vspace{.25cm}
    \caption{Actual example taken from the dataset with multiple SDR panorama viewpoints for the same instruction: \emph{Two parked bicycles, and a discarded couch, all on the left. Walk just past this couch, and stop before you pass another parked bicycle. This bike will be white and red, with a white seat. Touchdown is sitting on top of the bike seat.}}\label{fig:sdr}
\end{figure*}

To address these challenges, we collect, check and release the Touchdown panoramas as part of an update to the 114k existing StreetLearn panoramas, which cover regions of New York City and Pittsburgh.
As shown in Figure \ref{fig:td_sl_overlay}, StreetLearn encompasses the entire region of New York City contained in Touchdown; however, the StreetLearn panoramas themselves are not sufficient for supporting the Touchdown tasks themselves. This is for several reasons. 

\begin{itemize}[nosep]
    \item The \textbf{granularity} of the panorama spacing is different. Figure \ref{fig:td_sl_overlay} shows that most of the panoramas are different. Touchdown has roughly 25\% of the panos but covers half of Manhattan compared to StreetLearn. 
    \item The language instructions refer to \textbf{transient objects} such as cars, bicycles, and couches, as illustrated in Figures \ref{fig:td_car_pano} and \ref{fig:sdr}. A panorama from a different time period will not contain these objects, so the instructions are not stable across time periods.
    \item Spatial description resolution requires coverage of multiple points-of-view for those specific panoramas. Figure \ref{fig:sdr} shows an example SDR description and the corresponding views from which it can be answered.
\end{itemize}

In all, the Touchdown tasks encompass 29,641 panoramas. All of these went through extensive manual review by annotators to check for personally identifiable information (PII), such as faces and license plates. Regions containing PII were marked as bounding boxes by annotators, and we blurred all of these regions for the final images.

\begin{table*}
\centering
 \begin{tabular}{lcccc} 
 \textbf{Method} & \textbf{A@40px} $\uparrow$ & \textbf{A@80px} $\uparrow$ & \textbf{A@120px} $\uparrow$ & \textbf{Dist} $\downarrow$ \\
 \hline
\addlinespace[0.2cm]
 \multicolumn{5}{l}{\textit{Development}}\\
 \cite{Chen19:touchdown} & 24.81 & 32.83 & 36.44 & 729\\ 
 Retouch-\textsc{LingUNet} & 29.79 & 35.28 & 38.14 & 800 \\
\addlinespace[0.2cm]
 \multicolumn{5}{l}{\textit{Test}}\\
 \cite{Chen19:touchdown} & 26.11 & 34.59 & 37.81 & 708\\ 
 Retouch-\textsc{LingUNet} & 30.32 & 36.73 & 39.27 & 793 \\[.2cm]
\end{tabular}
\caption{SDR development and test results using the \textsc{LingUNet} architecture, which  \citet{Chen19:touchdown} reported as the best performing system. }
\label{table:sdr}
\end{table*}

\begin{table*}
\centering
 \begin{tabular}{lrrrrr} 
 \textbf{Method} & \textbf{TC} $\uparrow$ & \textbf{SPD} $\downarrow$ & \textbf{SED} $\uparrow$ & \textbf{nDTW} $\uparrow$ & \textbf{SDTW} $\uparrow$ \\ 
\hline
\addlinespace[0.2cm]
 \multicolumn{6}{l}{\textit{Development}}\\
 \cite{Chen19:touchdown} & 9.8 & 19.1 & 0.094 & &\\ 
 Retouch-\textsc{RConcat} & 13.4 & 17.1 & 0.124 & 4.9 & 1.3 \\ 
\addlinespace[0.2cm]
 \textit{Test}\\
 \cite{Chen19:touchdown} & 10.7 & 19.5 & 0.104 & &\\ 
 Retouch-\textsc{RConcat} & 12.8 & 17.1 & 0.131 & 5.0 & 1.4 \\[.2cm]
\end{tabular}
\caption{Navigation development and test results. We use the \textsc{RConcat} architecture, which \citet{Chen19:touchdown} reported as the best performing.}
\label{table:navigation}
\end{table*}

\section{Experiments}
\label{sec:exp}

We re-implement the best-reported models on the navigation and spatial description resolution tasks from \cite{Chen19:touchdown} to compare performance with our data release to the original Touchdown paper. 
The key difference between the two settings is that our released panoramas contain additional blurred patches (Section~\ref{sec:process}). Another minor difference is that we use a word-piece tokenizer~\citep{bert} instead of a full-word tokenizer. 

\paragraph{Spatial Description Resolution.}
SDR results are given in Table \ref{table:sdr}. Following \cite{Chen19:touchdown}, we report mean distance error and accuracy with different thresholds (40px, 80px, and 120px), which measures the proportion of evaluation items where the pixel chosen by the model is within the specified pixel distance. Our Retouchdown reimplementation of \textsc{LingUNet} obtains better performance on the accuracy measures, but worse performance on mean distance error. To check whether this is a consequence of the blurring, we ran our model with features retrieved from original panoramas and obtained similar results as those listed in Table \ref{table:sdr}. Given this, the performance difference between our model and the original paper are likely not due to the additional blurring. As such, the Touchdown panoramas available through StreetLearn can be reliably used as direct replacement for those used in \cite{Chen19:touchdown}.

\paragraph{Vision-and-Language Navigation.} We use the following metrics to evaluate VLN performance:

\begin{itemize}[nosep]
\item Task Completion (TC): the accuracy of navigating to the correct location. The correct location is defined as the exact goal panorama or one of its neighboring panoramas. This is the equivalent of the success rate metric (SR) used commonly in VLN for R2R.
\item Shortest-path distance (SPD): the mean of the distances over all executions of the agent's final panorama position and the goal panorama.
\item Success weighted by Edit Distance (SED): normalized graph edit distance between the agent path and true path, with points only awarded for successful paths.
\item Normalized Dynamic Time Warping (nDTW): a minimized cumulative distance between the agent path and true path, normalized by path length.
\item Success weighted Dynamic Time Warping (SDTW): nDTW, with points awarded only for successful paths.
\end{itemize}

\noindent
TC, SPD, and SED are defined in \cite{Chen19:touchdown} and nDTW and SDTW are defined in \cite{ilharco:etal:dtw}.

VLN results are given in Table \ref{table:navigation}. 
Our Retouchdown reimplementation of the \textsc{RConcat} model improves over the results given in \cite{Chen19:touchdown} for all metrics. We also establish benchmark scores for nDTW and SDTW. As with SDR, the panoramas now available via StreetLearn thus do not remove information critical for the VLN task. In our implementation, we use imitation learning on top of a scalable framework based on the Actor-Learner architecture \citep{lansing2019valan}, instead of supervised learning using Hogwild!~\citep{hogwild}.
These differences likely explain the observed differences with the original results. 

Compared to interior navigation in the Room-to-Room (R2R) task, the Touchdown task is much harder: e.g. the current state-of-the-art success rate (equivalent to TC) for R2R on the validation unseen dataset is 55\% \citep{zhu2019visionlanguage}. The same holds for DTW measures: \cite{ilharco:etal:dtw} report a success rate of 44\% and corresponding SDTW of 38.3\% for a fidelity-oriented version of the Reinforced Cross-modal Matching agent \citep{Wang:2018:RCM}. The TC of 12.8\% and SDTW of 1.4\% obtained by Retouch-\textsc{RConcat} amply demonstrates the challenge of the outdoor navigation problem defined by Touchdown. The greater diversity of the visual environments and the far greater degrees-of-freedom for navigation thus provide plenty of headroom for future research.

\section{Conclusion}

The research community is interested in using large-scale resources such as Street View for work on computer vision and navigation. In order to comply with Street View's terms-of-service (which allow for only limited use of its data and APIs) and with its data restrictions, we have enriched StreetLearn with panoramas from the Touchdown study. That dataset is periodically updated to comply with Google Street View takedown requests to respect individuals' privacy preferences. We encourage the research community to use only vetted and approved resources like StreetLearn, including our new release of the Touchdown panoramas, for their Street View oriented work.

The addition of Touchdown to StreetLearn (a.k.a. \textit{Retouchdown}) boosts the total panorama count for the StreetLearn dataset\footnote{\url{http://streetlearn.cc}} from 114k to 144k. Furthermore, it contains multiple panoramas from the same neighborhoods, which supports work on learning to navigate in a region and testing in that same region using panoramas from a different time. Our code for training and evaluating vision-and-language navigation agents and spatial description resolution models are publicly available as part of the VALAN framework \citep{lansing2019valan}.\footnote{Code: \url{https://github.com/google-research/valan}}



\section*{Acknowledgments}

The authors would like to thank Howard Chen for his assistance as we reproduced the Touchdown results, Larry Lansing, Valts Blukis and Vihan Jain for their help with the code and open-sourcing, the Google Maps and Google Street View teams for their support in accessing and releasing the data, and the Google Data Compute team for their help with panorama review and blurring.

\bibliography{vln}
\bibliographystyle{apalike}

\end{document}